\documentclass{llncs}

\usepackage[utf8]{inputenc}
\usepackage{amsmath}
\usepackage{array}
\usepackage{tabularx}
\usepackage{multirow}
\usepackage{graphicx}
\usepackage{amsfonts}
\usepackage{amssymb}
\usepackage{url}
\usepackage{tcolorbox}
\usepackage{enumitem}
\usepackage{graphicx}
\usepackage{multirow, makecell}
\usepackage{lingmacros}
\usepackage{bbm}
\usepackage{booktabs}
\usepackage[T1]{fontenc}% http://ctan.org/pkg/fontenc
\usepackage[outline]{contour}% http://ctan.org/pkg/contour
\usepackage{xcolor}% http://ctan.org/pkg/xcolor
\contourlength{0.2pt}
\contournumber{55}%

\usepackage{cite} 
\usepackage{hyperref}

\newcolumntype{C}{>{\centering\arraybackslash}X}
\newcommand{\heading}[1]{\multicolumn{1}{c|}{#1}}

\begin{document}

{\let\thefootnote\relax\footnotetext{Copyright \textcopyright\ 2020 for this paper by its authors. Use permitted under Creative Commons License Attribution 4.0 International (CC BY 4.0). CLEF 2020, 22-25 September 2020, Thessaloniki, Greece.}}

% \title{QMUL-SDS at \texttt{CheckThat!} 2020:\\ Determining COVID-19 Tweet Check-Worthiness}
\title{QMUL-SDS at \texttt{CheckThat!} 2020:\\ Determining COVID-19 Tweet Check-Worthiness Using an Enhanced CT-BERT with Numeric Expressions}

\author{Rabab Alkhalifa\inst{1,5}, Theodore Yoong\inst{4}, Elena Kochkina\inst{2,3}, Arkaitz Zubiaga\inst{1}, \and Maria Liakata\inst{1,2,3}}
\institute{
Queen Mary University of London, United Kingdom
% \url{jane@university.it}
\and
University of Warwick, United Kingdom
%\url{john@university.de} 
\and
Alan Turing Institute, United Kingdom
% \url{john@university.de} 
\and
University of Oxford, United Kingdom
%\url{john@university.de} 
\and
Imam Abdulrahman bin Faisal University, Saudi Arabia
}
\maketitle

\begin{abstract}
 This paper describes the participation of the QMUL-SDS team for Task 1 of the CLEF 2020 \texttt{CheckThat!} shared task. The purpose of this task is to determine the check-worthiness of tweets about COVID-19 to identify and prioritise tweets that need fact-checking. The overarching aim is to further support  ongoing efforts to protect the public from fake news and help people find reliable information. We describe and analyse the results of our submissions. We show that a CNN using COVID-Twitter-BERT (CT-BERT) enhanced with numeric expressions can effectively boost performance from baseline results. We also show results of training data augmentation with rumours on other topics.  Our best system ranked fourth in the task with encouraging outcomes showing potential for improved results in the future.
\end{abstract}

\section{Introduction}
%Although seeking health advice from medical professionals remains extremely important, 
The vast majority of people seek information online and consider it a touchstone of guidance and authority \cite{miller2012online}. In particular, social media has become the key resource to go to for following updates during times of crisis \cite{palen2008online}. Any registered user can share posts on social media without content verification, potentially exposing thousands or millions of other users to harmful misinformation. To prevent the undesired consequences of misinformation spread, there is a need to develop tools to assess the validity of social media posts. This problem has been particularly accentuated in light of the COVID-19 pandemic, accompanied by the rising spread of unverified claims and conspiracy theories about the virus and untested dangerous treatments. Compounded with the devastating effects from the virus alone, the social harms from misinformation spread can be particularly injurious \cite{cuan2020misinformation}. The \texttt{CheckThat!} shared task provided a benchmark evaluation lab to develop systems for check-worthiness detection, with the aim of prioritising claims to be provided to fact-checkers.

% Managing the flood of information poses various challenges, both to social media platforms and to users who are exposed to it, given that doing so manually is not feasible. From a natural language processing (NLP) perspective, an automated approach to minimise the transmission of false information is to understand the contextual information spread and prioritise the verification process \cite{zubiaga2018detection}.

% Fact-checking organisations play an important role here, professionally verifying and disseminating verdicts on selected pieces of information that circulate online and in other media. One of the areas where fact-checkers seek support of technology is in the identification of key claims to be checked \cite{babakar2016state}, also referred to as claim detection or check-worthiness determination. Automated identification and prioritisation of claims to be verified can save to fact-checkers who can instead dedicate their time to checking the veracity of those claims. This makes the task determination of check-worthiness a paramount task, which we focus on here.

In this paper, we present our approaches in tackling the check-worthiness detection task as outlined in Task 1 of the CLEF-2020 \texttt{CheckThat!} Lab. We evaluated several variants of our Convolutional Neural Network (CNN) model with different pre-processing approaches and several BERT embeddings. We also tested the benefits of including the use of external data to augment the training data provided. We submitted three models that have shown the best performance on the development set. Our best performing model utilised a COVID-Twitter-BERT (CT-BERT) enhanced with numeric expressions, which was ranked fourth in the task.

\section{Related Work}
\vspace{20pt}
\label{sec:related}
We organise the related work into two subsections relevant to our proposed methods and the systems we submitted to the evaluation lab: claim check-worthiness and rumour detection.% and transformer models.

\subsection{Determination of Claim Check-worthiness}

While there is no general streamlined approach to fact-checking, the fact-checking pipeline can be divided into different sub-tasks based on a number of contexts \cite{babakar2016state}. The first of the tasks and the one concerning our work consists in producing a list of claims ranked by importance (check-worthiness), in an effort to prioritise claims to be fact-checked. Systems for claim detection (as a classification task) and ranking by check-worthiness include (i) ClaimBuster \cite{hassan2017claimbuster}, which combines numerous features such as TF-IDF, POS tags and NER on a Support Vector Machine to produce importance scores for each claim, and (ii) ClaimRank \cite{Gencheva2017ACA}, which uses a large set of features both from individuals sentences and from surrounding context. More recent methods, such as \cite{Konstantinovskiy2018}, have made use of embedding-based methods such as InferSent for detecting claims by leveraging contextual features within sentences. In previous editions of CheckThat! \cite{hasanain2019overview}, the shared task did not involve the detection of claims, as claims were already given as input.

Previous work on claim detection by Konstantinovskiy et al. \cite{Konstantinovskiy2018} suggested the use of numeric expressions as a strong baseline for detection of claims. Indeed they showed that the use of numeric expressions as a feature leads to high precision, despite achieving lower recall and overall F1 score than other methods. This is due to the prevailing presence of numeric expressions in check-worthy claims, as opposed to non-check-worthy claims and non-claims. Given the emphasis of the CheckThat! shared task on precision-based evaluation (using mean average precision as a metric), we opted for incorporating numeric expressions in our model.
%Next, counting the nuances of utterance itself involve determining what makes statements check-worthy by analysis the user's profile (veracity checking). Then, the final process of fact-checking where the task is to infer the conformity of each check-worthy sentence based on the contextual information and organisation policy.
% The process of automatic prioritisation of claims is challenging and it is the main contribution of this paper. Check-worthiness is a task which involves highlighting suspicious information, making it easier for people to judge the veracity of authoritative information in social media platforms.

\subsection{Rumour Detection}

A rumour is generally defined as an unverified piece of information that circulates. In the same way that a check-worthiness detection looks at claims to be verified, e.g. in the context of a TV debate, rumour detection consists in detecting pieces of information that are in circulation while they still lack verification, generally in the context of breaking news, making it a time-sensitive task \cite{zubiaga2018detection}. Rumours differ from check-worthy claims in their nature as well as relevance to the fact-checkers, as not all rumours are necessarily of interest to fact-checkers. Still, both tasks have significant commonalities.

In our approaches to check-worthiness determination, we try to leverage existing data for rumour detection, consisting of rumours and non-rumours, with the aim of providing additional knowledge that would enrich the task (see \S \ref{ssec:corpora}).

Rumour detection, as the task of detecting unverified pieces of information, has been studied before, for instance through the RumourEval shared tasks held at SemEval 2019 \cite{gorrell2019semeval}. Prior to that, Zubiaga et al. \cite{zubiaga2017exploiting} introduced a sequential rumour detection model that leveraged Conditional Random Fields (CRF) for leveraging event context, as well as Zhao et al. \cite{zhao2015enquiring} that looked at evidence from others responding to tweets with comments of the form of \textit{``is this really true?''}, which would be indicative of a tweet containing rumourous content.

\section{Task Description}
\label{sec:task}
The task we explore was introduced by  Barr{\'o}n-Cede{\~n}o et al. \cite{clef2020original} and is formulated as follows:

\vspace{10pt}
\vskip1.5truemm
\begin{tcolorbox}[colback=red!5!white,colframe=red!75!black]
Given a topic and a stream of potentially-related tweets, rank the tweets according to their check-worthiness for the topic, where a check-worthy tweet is a tweet that includes a claim that is of interest to a large audience (especially journalists) and may have a harmful effect.
\end{tcolorbox}
\vskip1.5truemm
\vspace{10pt}
\noindent For example, consider the target topic–tweet pair:

\vspace{10pt}
\vskip1.5truemm
\begin{tcolorbox}[colback=yellow!5!white,colframe=yellow!75!black]
\textbf{Target topic}: COVID-19
\vskip1.5truemm
\textbf{Tweet}: Doctors in \#Italy warn Europe to “get ready” for \#coronavirus, saying ~10\% of \#COVID19 patients need ICU care, and hospitals are overwhelmed. \vskip1.5truemm
\textbf{Label}: Check-worthy 
\end{tcolorbox}
\vskip1.5truemm
\vspace{10pt}

Although Task 1 is available in both English and Arabic, we focussed solely on the English task \cite{clef-checkthat-lncs:2020}. Tweets in this dataset for this task exclusively covered the pandemic caused by the Coronavirus Disease 2019 (COVID-19). The ultimate objective of ranking the tweets identified as check-worthy claims is to enable prioritisation of claims to fact-checkers.

The task can be formally described as the following binary classification problem. We define the training set consisting of $n$ labelled tweets as $\mathcal{D}=\{({\bf x}_i,y_i),1\leq i\leq n\}\in\left(\mathcal{X}\times\{0,1\}\right)^n$. Here, ${\bf x}_i$ is the $i^{\text{th}}$ feature vector in feature space $\mathcal{X}$ which contains the tweet features such as the $i^{\text{th}}$ tweet itself ${\bf t}_i$, the topic, and whether it is a claim or not; and $y_i\in\{0,1\}$ is the label indicating check-worthiness of ${\bf t}_i$. The objective is to obtain a map $h:\mathcal{X}\mapsto\{0,1\}$, based on the class probability measure $\mathbb{P}(y|{\bf t})$, which is subsequently used to rank the tweets in the test set. %and $p_i$ is a class probability, which we subsequently use to rank all the items in the test set. %reference organiser's papers

\subsection{Datasets}
\label{ssec:corpora}

\iffalse
\begin{table}[t]
\centering
\normalsize
%\resizebox{\textwidth}{!}{%
\begin{tabular}{|l|l|l|l|}
\hline
                 & \begin{tabular}[c]{@{}l@{}}N tweets\\Check-worthy\\ (rumours)\end{tabular} & \begin{tabular}[c]{@{}l@{}}N tweets\\ Not check-worthy\\ (non-rumours)\end{tabular} & Total \\ \hline
CLEF Train & 231 & 441 & 672 \\ \hline
CLEF Development & 59 & 91 & 150 \\ \hline
CLEF Test & 80 & 60 & 140 \\ \hline
PHEME & 2402 & 4023 & 6425 \\ \hline
Twitter 15 & 1012 & 362 & 1374 \\ \hline
Twitter 16 & 536 & 199 & 735 \\ \hline
\end{tabular}%
%}
\caption{Number of posts and class distribution in the datasets used}
\label{tab:datasets}
\end{table}
\fi

%\begin{center}
\begin{table}[htb]
\centering
% \scriptsize
{\renewcommand{\arraystretch}{2}
\begin{tabular}{|c|c|c|c|}
\hline
$\quad${Dataset}$\quad$ & {\renewcommand{\arraystretch}{1}\begin{tabular}[c]{@{}c@{}}\tiny{$\quad$}\\{No. of check-}\\$\quad${worthy tweets}$\quad$\\{(rumours)}\\\tiny{$\quad$}\end{tabular}} & {\renewcommand{\arraystretch}{1}\begin{tabular}[c]{@{}c@{}}$\quad${No. of non-check-}$\quad$\\{worthy tweets}\\{(non-rumours)}\end{tabular}} & $\quad${Total}$\quad$ \\ \hline
CLEF Train & 231 & 441 & 672 \\ \hline
CLEF Development & 59 & 91 & 150 \\ \hline
CLEF Test & 80 & 60 & 140 \\ \hline
PHEME & 2402 & 4023 & 6425 \\ \hline
Twitter 15 & 1012 & 362 & 1374 \\ \hline
Twitter 16 & 536 & 199 & 735 \\ \hline
\end{tabular}%
}
\normalsize
\vskip3truemm
\caption{Number of posts and class distribution in the datasets used}
\label{tab:datasets}
\end{table}
%\end{center}
In our experiments, we made use of three datasets of Twitter posts in English, which include the dataset provided by the organisers (CLEF) and two external publicly available datasets (PHEME, Twitter 15 and Twitter 16) to augment the training set. The PHEME, Twitter 15 and Twitter 16 datasets were chosen for augmentation as these are  relatively large datasets annotated for rumour detection task, which is very similar to claim check-worthiness as described in section \ref{sec:related}. Table~\ref{tab:datasets} shows the number of tweets used in each of the datasets and the class distribution. 

The \emph{CLEF} dataset contains tweets related to the topic of COVID-19. They were annotated by the task organisers as either check-worthy or not check-worthy thus defining a binary classification task.\footnote{Annotation rules can be found here: \url{https://github.com/sshaar/clef2020-factchecking-task1\#data-annotation-process}} 
This dataset is rather small and is limited to individual tweets concerned with a single topic. The dataset is  imbalanced with the majority of tweets being not check-worthy.

% frequency of hashtags, urls, digits and accounts
%figure  Table x. Example of Check-Worthiness
% -Training data analysis
% Only XXXX were labelled as check-worthy– a only imbalance with only ??? of the dataset are of the same topic. 
% -Testing data analysis
% -Evaluation data analysis

The \emph{PHEME} dataset~\cite{zubiaga2018detection, kochkina2018all} contains Twitter conversations discussing rumours (defined as unverified check-worthy claims spreading widely on social media) and non-rumours.\footnote{\url{https://figshare.com/articles/PHEME_dataset_for_Rumour_Detection_and_Veracity_Classification/6392078}}  This dataset contains conversations related to 9 major newsworthy events, such as shooting in Charlie Hebdo, shooting in Ottowa, crash of Germanwings plane. In this work, we use only the source tweets of the conversations in the PHEME dataset (they are conveying the essence of a rumour, rather than the following discussion) in order to have the same input structure as the CLEF dataset. We performed experiments augmenting the training set with both rumours and non-rumours from the PHEME dataset. We found that adding rumours only is more beneficial than adding the full PHEME dataset. 

The \emph{Twitter 15 and Twitter 16} datasets~\cite{ma2017detect} contain Twitter conversations, discussing True, False and Unverified rumours as well as non-rumours on various topics. Here, we do not use all 4-class labels, but instead convert True, False and Unverified classes into single check-worthy class. We also use only source tweets to augment the CLEF training set. 

\subsection{Evaluation}
The CLEF dataset is split into training, development and testing sets. The check-worthiness task is evaluated as a ranking task, i.e. the participant systems should produce a list of tweets with the estimated score for check-worthiness. The official evaluation metric is Mean Average Precision (MAP), but the precisions at rank $k$ ($P@5,P@10,P@30$) are also reported. Baseline results provided by the organisers are Random Classification (MAP = 0.35) and SVM with $N$-gram Prediction (MAP = 0.69).

% \vspace{20pt}
\section{Our Approach}
% \vspace{20pt}
\label{sec:approach}

%While frequency-based classical models (e.g. with support vector machines \cite{Joachims1998}) have been successful, deep learning models are slowly becoming more commonplace in NLP due to their quick and accurate training and prediction results across a plethora of classification tasks. Vector representations in deep learning by the mean of converting sentence tokens into a vector, forming a matrix for feature representation from word distribution and sentences context has been widely used.\\

In our approach, we fed a pre-trained word-level vector representation into a CNN model. Using vector representations of words as inputs offers high flexibility, allowing swaps between different pre-trained word vectors during model initialisation to be maintained without additional overhead. For our work, we tested multiple feature representations in order to assess which one would be best for our problem, these include combinations of frequency-based, vector-based representations and authors' profiles. Since most evaluation tweets were not supplied with author's profiles, and frequency-based models may not be of great generalisability for unseen data, we reduced our exploration to two dynamic word embeddings, ELMo \cite{peters2018deep} and BERT \cite{devlin2019bert}. Since the former did not provide any increase in performance, we chose the latter to be further explored with different pre-processing techniques.

\vspace{20pt}
\begin{figure}
 \begin{center}
  \includegraphics[width=1\textwidth]{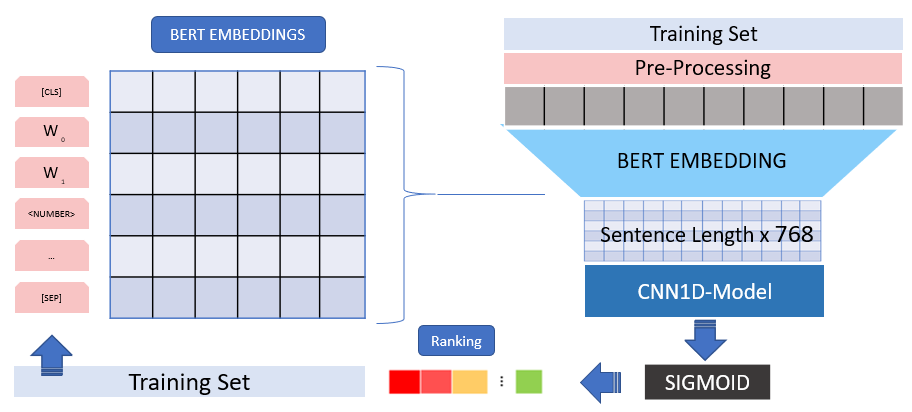}
  \caption{General Model Architecture.}
  \label{fig:general-model}
 \end{center}
\end{figure}

%\vspace{20pt}
With these settings, we evaluated our model using two variations of the BERT pre-trained architecture, uncased BERT (uncased-BERT) and COVID-Twitter-BERT (CT-BERT) \cite{covidBERT} (see Figure \ref{fig:general-model}). While both are transformer-based models, CT-BERT is pre-trained on COVID-19 Twitter data using whole word-masked modelling and next sentence prediction.

In the following sections we describe the main characteristics of our designed system including pre-processing, feature representation, model architecture and hyperparameters used.

\subsection{Pre-processing}

We performed standard Twitter data pre-processing in order to improve our system performance. For each model, we implemented different pre-processing variants depending on the vector representation leading to best performance. The pre-processing steps can be summarised as follows.

\begin{enumerate}[label=(\roman*)]
\item \textbf{Segment2Token}: We split each sentence into tokens considering the type of every segment and breaking it into individual tokens. Digits, URLs, accounts and hashtags were replaced by $\langle number\rangle$ , $\langle url\rangle$, $\langle account\rangle$, $\langle hashtag\rangle$. respectively. This was implemented using a simple split function with different expression finding methods. By analysing generated segments, we settled on different treatment for special tokens in every tweet. For example, \textbf{hyperlinks} were either completely removed from the dataset or replaced by special tokens. Furthermore, \textbf{digits} and all other numerical expressions that contained `\%' or `\$' were either removed or tokenised. Tokenising numerical expressions allows the model to generalise better (see \S \ref{sec:results}). For example:
\vskip1.5truemm
\begin{tcolorbox}[colback=yellow!5!white,colframe=yellow!75!black]
\textbf{Tweet}: $[NEWS]$ Naver \#BAEKHYUN EXO Baekhyun donates 50 million won to prevent the spread of Corona 19 {@}weareoneEXO \#EXO
\vskip1.5truemm
\textbf{Segment2Token}: $[NEWS]$ Naver $\langle hashtag\rangle$ EXO Baekhyun donates $\langle number\rangle$ won to prevent the spread of Corona 19 $\langle account\rangle$ $\langle hashtag\rangle$.
\end{tcolorbox}
\vskip1.5truemm
\vspace{10pt}

\item \textbf{Segment2Root}:
In NLP, the $\tilde{\chi}^2$-statistical measure tests term-dependency of the tweet being about one of the classes as in \cite{olteanu2014crisislex}. We used it to analyse the segments of the tweets. In these settings, for few account handles and hashtags with high $\tilde{\chi}^2$-score, we manually combined them depending on their semantic meaning. For instance:
\vspace{5pt}
\vskip1.5truemm
\begin{tcolorbox}[colback=yellow!5!white,colframe=yellow!75!black]
\textbf{Hashtags}: \#coronavirus, \#COVID19', \#COVID-19, \#COVID19, \#Coronavirus, \#Corona-virus
\vskip1.5truemm
\textbf{Hashtag2Root}: coronavirus
\end{tcolorbox}
\vskip1.5truemm
\vspace{5pt}
In this case, different hashtags about COVID-19 were all consolidated under the `coronavirus' umbrella term.\\
% Further, we find out that some hashtags with similar meaning are negatively correlated (check heatmap)
% \begin{figure}
%  \begin{center}
%   \includegraphics[width=.7\textwidth]{imgs/hashtag_corelation.png}
%   \caption{Hashtags}
%   \label{fig:evaluation-table}
%  \end{center}
% \end{figure}
% \vspace{20pt}

% - using aspect to understand context was implemented (figure), however this most likely requiring more training data than was provided in this competition did not have any improvement when implemented 
% Lemmatisation?no
\item \textbf{Word2id}: As with all BERT models, we included classification embedding tokens for every tweet: \texttt{[CLS]} at the beginning and a separator token \texttt{[SEP]} at the end. We then decomposed ${\bf t}_i$ into a sequence of numerical tokens using BERT tokenisation methods. This was done by mapping each token to a unique integer in the corpus’ vocabulary.
\vspace{5pt}
\item \textbf{Padding}: We ensured that the input sequences in every batch was the same length. This was achieved through increasing the length of some of the sequences by adding more tokens. We tried to reduce the padding by allowing our model to decide the padding length based on a given batch size (set to 10) and the longest sequence within the given batch. For example, if the longest sequence length for a given batch is 20, then all other shorter sentences will be padded to match its length.
\vspace{5pt}
\end{enumerate}

Finally, a look-up table was used for each token from the generated representation, ready to be fed into the model.

\subsection{Model Hyperparameters}
\vspace{5pt}
The CNN architecture requires tuning various hyperparameters. These include input representations, number of layers and filters, pooling, and activation functions. We utilised a BERT language model in accompaniment to the the general CNN architecture. Within the model design, we propose variations of a three-layer CNN with 32 filters with different window sizes: 2, 4 and 7. The multiple filters act as feature extractors.

Additionally, we used an Adam optimiser with learning rate fixed at $2\text{e}^{-5}$ and number of training epochs set to 8. Our $N$-gram kernels encompass a Rectified Linear Unit (ReLU) activation function, given by $\max(0,x)$. All pooling layers use a max-pooling operation. For the binary classification, we utilise a sigmoid activation function $\sigma(x)$ for the output layer, defined as
$$\sigma(x)=\frac{1}{1+\text{e}^{-x}}$$
To determine the final output labels, we classify check-worthiness based on the indicator variable
\[h(x)=\begin{cases}
1\quad\text{if }\sigma(x)\geq\displaystyle\frac12\text{, indicating a check-worthy tweet,}\\
\\
0\quad\text{if }\sigma(x)<\displaystyle\frac12\text{, suppressing the tweet as non-check-worthy.}
\end{cases}\]
\section{Results and Discussion}
% \vspace{20pt}
\label{sec:results}

In the following section, we discuss the selection of the models we tried and ultimately submitted to the shared task. We also evaluate and compare their performance.

\subsection{Model Selection on the CLEF Development Set}
We performed our model selection using the development set. Details of the pre-processing steps and embeddings applied to each of the eight models we tested are given in Table \ref{tab:models}. The performance of the models according to the various precision metrics are shown in Table \ref{tab:evaluation-table}.

\subsubsection{CLEF Benchmark Data Experiments:} Text distortion has been used by \cite{ghanem2018upv}, where their methods were more successful than using the full text in the classification process. Taking inspiration from their work, we used tokenisation for different segments of the tweet where account handles, hashtags, URLs and digits were assigned special tokens and added to the model vocabulary. The goal of this step was to avoid over-fitting the training data.

In Model 7, we experimented with ELMo embeddings, which gave the best performance in terms of MAP, in our tests without additional pre-processing and outperforms random baseline (MAP = 0.35). However, Model 7 did not outperform the $N$-gram baseline (MAP = 0.69), and thus we did not choose it for the test set submission.

In Model 4, we trained word embeddings on the training set along with the model and combined them with TF-IDF representations. This led to improvements over Models 5-7 and over the $N$-gram baseline.

For Model 1, we used intensive pre-processing in tandem with a CNN model with three filters of sizes 2, 4 and 7. On the other hand, in Model 2, only numeric expressions were tokenised and the CNN model only had two filters of sizes 2 and 4. Models 1 and 2 displayed the best performance on the development set, and were hence chosen for submission on the testing set.

\begin{center}
\begin{table}[htb]
{\renewcommand{\arraystretch}{2}
\begin{tabularx}{\textwidth}{|c|X|c|}%
\hline
\multicolumn{1}{|c|}{Model No.} & \heading{Pre-processing} & \heading{Embeddings} \\ \hline
1 & Frequently mentioned entities replaced with their account name. Hashtags with repeated topics combined using the $\tilde{\chi}^2$-score, other URLs and hashtags tokenised with special tokens. & CT-BERT \\ \hline
2 & Special tokens for digits. Account handles, URLs and hashtags removed. & CT-BERT \\ \hline
3 & Training set merged with rumours from PHEME.  Special tokens for digits. & \makecell[t]{BERT-EN \\ (Uncased)} \\ \hline
4 & Digits, account handles, URLs and hashtags removed. & \makecell[t]{CLEF Train Embeddings \\ + TF-IDF} \\ \hline
5 & Training set merged with Twitter 15 and Twitter 16 datasets. Special tokens for digits. & \makecell[t]{BERT-EN \\ (Uncased)} \\ \hline
6 & Training set merged with PHEME, Twitter 15 and Twitter 16 datasets. Special tokens for digits. & \makecell[t]{BERT-EN \\ (Uncased)} \\ \hline
7 & No pre-processing was applied. & ELMo \\ \hline
8 & Trained using PHEME, Twitter 15 and Twitter 16 datasets only, \textbf{without} CLEF training data. & \makecell[t]{BERT-EN \\ (Uncased)} \\ \hline
\end{tabularx}
}
\vskip1.5truemm
\caption{Description of the models tested.}
\label{tab:models}
\end{table}
\end{center}

\subsubsection{Data Augmentation Experiments:}
While the task definition states the presence of the target topic when identifying tweet check-worthiness, the dataset provided only covers a single topic, COVID-19. We experimented with training data augmentation using check-worthy tweets from external datasets (PHEME and Twitter15, Twitter 16 as described in section \ref{ssec:corpora}) (see models 3, 5, 6 in Table \ref{tab:models}). These datasets cover different topics, accounts and vocabulary, so incorporating them could contribute to future generalisability of the model. We also performed experiments using only existing external datasets of rumours and non-rumours and omitting the CLEF training data to test the generalisability of the currently available datasets and models to the emergence of new rumour topics (see model 8 in Table \ref{tab:models}). The results are presented in Table \ref{tab:evaluation-table}. 

\begin{center}
\begin{table}[ht]
% \scriptsize
{\renewcommand{\arraystretch}{2}
\begin{tabularx}{\textwidth}{|c|C|c|C|C|C|C|C|C|}
\hline
\multirow{2}{*}{\shortstack{Model\\No.}} & \multirow{2}{*}{\shortstack{Average\\precision}} & \multirow{2}{*}{\shortstack{\emph{R}-precision\\\textbf{($R=59$)}}} & \multicolumn{6}{c|}{Precision@\emph{k}} \\ \cline{4-9} 
 &  &  & @1 & @3 & @5 & @10 & @20 & @50 \\ \hline
1 & 0.81 & 0.71 & 1.00 & 1.00 & 1.00 & 1.00 & 0.95 & 0.74 \\ \hline
2 & 0.80 & 0.71 & 1.00 & 1.00 & 1.00 & 1.00 & 0.95 & 0.76 \\ \hline
3 & 0.75 & 0.69 & 1.00 & 0.67 & 0.60 & 0.80 & 0.85 & 0.74 \\ \hline
4 & 0.74 & 0.63 & 1.00 & 1.00 & 0.80 & 0.90 & 0.85 & 0.68 \\ \hline
5 & 0.65 & 0.59 & 1.00 & 1.00 & 0.80 & 0.80 & 0.75 & 0.62 \\ \hline
6 & 0.56 & 0.57 & 0.00 & 0.33 & 0.20 & 0.50 & 0.70 & 0.58 \\ \hline
7 & 0.53 & 0.56 & 0.00 & 0.33 & 0.40 & 0.50 & 0.55 & 0.58 \\ \hline
8 & 0.45 & 0.44 & 1.00 & 0.33 & 0.40 & 0.40 & 0.50 & 0.44 \\ \hline
\end{tabularx}%
}
\normalsize
\vskip5truemm
\caption{Model performance on the development set}
\label{tab:evaluation-table}
\end{table}
\end{center}

%Table x. MAP scores for each model across the provided datasets [ Train, eval]
%Table x. MAP scores for each model across labels [test]
% [Model - Scores]
% TP,TN,FP,FN 

As expected, Model 8, which did not use the CLEF training data, performed worse compared to the models that did make use of the training data provided. However, it outperformed the random baseline (MAP = 0.35) by 10\%, showing that there is enough overlap in the task definitions and inherent nature of rumours/check-worthy claims to provide meaningful signal for model training.  Models 3, 5 and 6, which augmented the training data, did not perform as well as the models using only the training data. Models 5 and 6 did not outperform the $N$-gram baseline (MAP = 0.69) provided by the organisers\footnote{\url{https://github.com/sshaar/clef2020-factchecking-task1\#baseline}}. Model 3 only adds rumours to the training data, thus shifting the class balance in the dataset, and performed better than adding both rumours and non-rumours from PHEME to the training set.  

These results show the importance of model training or fine-tuning on the evaluation domain. The lack of performance improvement could be also due to the differences in the definitions of the tasks and rumours/check-worthy claims by each of the datasets. Moreover, different fact-checking organisations would naturally make different choices when analysing the same data. In \cite{Konstantinovskiy2018}, they found that educational background can lead to bias in annotation efforts for fact-checking. The subjectiveness that underlies check-worthiness thereby adds further complications to the task of ranking by importance. These results also highlight the especially challenging aspects of the need for generalising to new unseen topics, as well as leveraging data from a related task such as that of rumour detection, in detecting tweet check-worthiness.

\subsection{Results on the CLEF Test}
% - Experimental results are provided in Table
% - we use the benchmark training and test data provided in one experiment while we used more data in another one
% - Impact of the Features, embedding
% - Choice of Approach-Traditional Learning Over Deep Learning
% - best performances was by using covid-19 BERT, pre-process and tokenising numbers with special tokens. As the primary run we submitted our
We selected the best three models based on MAP (see Table \ref{tab:evaluation-table}). Table \ref{tab:testing-table} shows the official results obtained by our systems on the testing set.

\begin{center}
\begin{table}[htb]
%\scriptsize
{\renewcommand{\arraystretch}{2}
\begin{tabularx}{\textwidth}{|c|C|c|C|C|C|C|C|C|}
\hline
\multirow{2}{*}{\shortstack{Model\\No.}} & \multirow{2}{*}{\shortstack{Average\\precision}} & \multirow{2}{*}{\shortstack{$R$-precision\\($R=59$)}} & \multicolumn{6}{c|}{ Precision@\emph{k}} \\ \cline{4-9} 
 &  &  & @1 & @3 & @5 & @10 & @20 & @50 \\ \hline
1 & 0.71 & 0.63 & \textbf{1.00} & \textbf{1.00} & \textbf{1.00} & 0.90 & 0.80 & 0.64 \\ \hline
2 & \textbf{0.78} & \textbf{0.70} & \textbf{1.00} & \textbf{1.00} & \textbf{1.00} & \textbf{1.00} & \textbf{0.85} & \textbf{0.70} \\ \hline
3 & 0.73 & 0.63 & \textbf{1.00} & \textbf{1.00} & \textbf{1.00} & 0.90 & \textbf{0.85} & 0.68 \\ \hline
\end{tabularx}%
}
\normalsize
\vskip5truemm
\caption{Performance of the submitted models on the testing set}
\label{tab:testing-table}
\end{table}
\end{center}

\subsubsection{Models 1 and 2:}
We found that avoiding extensive tokenisation (Model 1) while merely tokenising numeric expressions yields better results in the test set and allows the model to learn more general patterns from the training set (Model 2). For example:
\vspace{5pt}
\vskip1.5truemm
\begin{tcolorbox}[colback=yellow!5!white,colframe=yellow!75!black]
\textbf{Tweet}: France, Spain and Germany are about 9 to 10 days behind Italy in $\#COVID19$ progression; the UK and the US follow at 13 to 16 days.
\vskip1.5truemm
\textbf{Digit2Token}: France, Spain and Germany are about $\langle number\rangle$ to $\langle number\rangle$ days behind Italy in corona virus progression; the UK and the US follow at $\langle number\rangle$ to $\langle number\rangle$ days.
\end{tcolorbox}
\vskip1.5truemm
\vspace{5pt}

Moreover, our approach uses a less complex model which keeps the weights small and results in better overall performance. Therefore, in order for our model to generalise to unseen tweets in the test set, numeric expressions should be unified and model complexity needs to be maintained. %higher resolution.

% Annotation Guideline: For each tweet, we would label it as a claim or not based on that definition. Some positive examples include: stating a definition, mentioning quantity in the present or the past, etc. Some negative examples include: spersonal opinions and preferences.

% proposed approach decreased generalisability due to extensive pre-processing
% - highest overall performance in the task was only MAP XX higher than our approach, making it ...
% \par
\subsubsection{Model 3:}
%PHEME ANALYSIS
In this submission, we augmented the training data with rumours from the PHEME dataset. While the results of this submission on the development set were the lowest out of the selected three, on the testing set it outperforms Model 1. This shows that external data from a related task adds meaningful signal for model training and contributes to system generalisability.

\vspace{5pt}
%Though, we confronted large performance drop compared to previous models due to the declined generalisability and density of the domain data, COVID-19 in this case. 

%This can be linked to many factors. Such as the difference between the training set and the PHEME dataset in terms vocabulary and coverage of topics. 

%\enumsentence{
%\textbf{Topic}:
%\newline
%\textbf{PHEME}: 
%}

% in the testing data proved to be non negligible

\section{Conclusion}
%Social media are becoming a non-negligible factor that influences the way in which the knowledge of readers is built. Nowadays, there is a need to filter what we are reading. This problem become a concern for platforms such as Twitter.\\

This paper describes our efforts as participants of the Task 1 of the \texttt{CheckThat!} 2020 evaluation lab, in which we ranked fourth, which was held in conjunction with the CLEF conference. We describe our proposed model that leverages the COVID-Twitter-19 BERT (CT-BERT) word embeddings and performs a special treatment for rare tokens with a CNN relying on the tweet alone.

The experimental results show that the performance of our model increases significantly by tokenising numerical expressions. The present work is restricted in choosing the best feature representation. In the future, this work can be enhanced in different possible directions. For example, incorporating pragmatic information related to author's profile information. Thus, simulate actual users' behaviour in verifying claims in social media.

Given the small size of the training data provided by the organisers, we also performed additional experiments leveraging external datasets with the aim of augmenting the training data. External data we incorporated had the challenge of being datasets pertaining to rumour detection and on different topics, hence with slight differences with the task and domain at hand. Our experiments with data augmentation did not lead to improved performance, highlighting that inclusion of external data of a different nature (i.e. in terms of task and domain) is particularly challenging and, if they can provide an improvement to the check-worthiness detection task, more careful integration and adaptation will be necessary.

\section{Acknowledgments}
This research utilised Queen Mary's Apocrita HPC facility, supported by QMUL Research-IT. http://doi.org/10.5281/zenodo.438045. This work was also partially supported by The Alan Turing Institute under the EPSRC grant EP/N510-129/1.

\bibliographystyle{IEEEtran}
\bibliography{bib}

\end{document}